\documentclass[10pt,twocolumn,letterpaper]{article}

\usepackage[applications]{wacv}             

\usepackage{microtype}
\usepackage{adjustbox}
\usepackage{multirow}
\usepackage{placeins}

\makeatletter
\renewcommand\paragraph{\@startsection{paragraph}{4}{\z@}%
  {1.0\baselineskip \@plus .2ex \@minus .2ex}%
  {-1em}%
  {\normalfont\normalsize\bfseries}}
\makeatother

\usepackage{dblfloatfix}

\definecolor{wacvblue}{rgb}{0.21,0.49,0.74}
\usepackage[breaklinks,colorlinks,allcolors=wacvblue]{hyperref}

\def\wacvPaperID{1474}
\def\confName{WACV}
\def\confYear{2027}

\title{Rethinking Densification in Gaussian SLAM: A Stationary Sampling Perspective}

\author{WeiTse Cheng \quad Yen-Jen Chiou \quad YuanFu Yang\\
National Yang Ming Chiao Tung University\\
{\tt\small yfyangd@nycu.edu.tw}
}

\makeatletter
\apptocmd{\@maketitle}{%
  \vspace{-6pt}%
  \begin{center}%
    \includegraphics[width=0.80\textwidth]{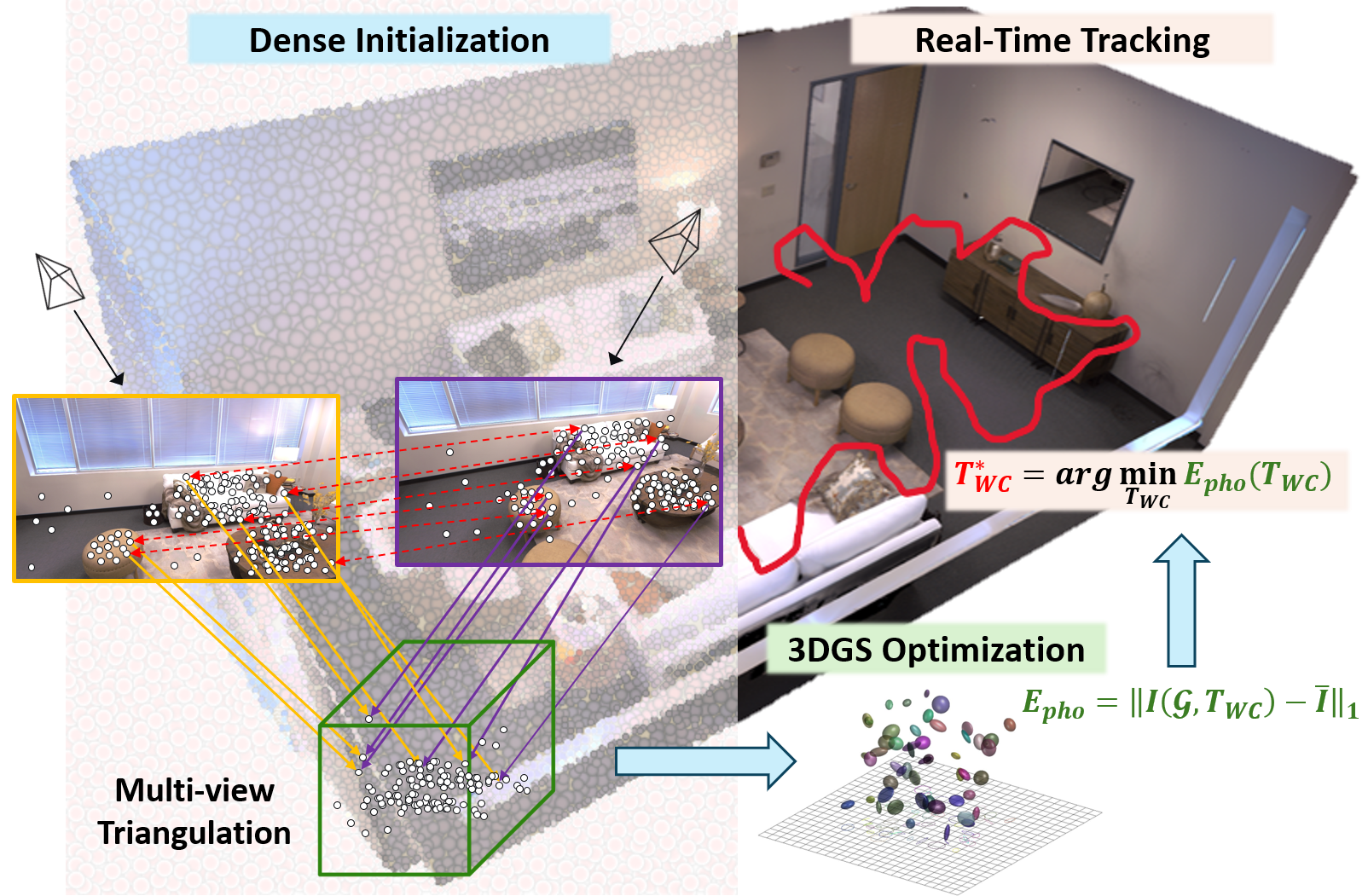}%
    \captionof{figure}{Overview of the proposed TopoGS-SLAM pipeline. The system integrates dense feature matching and multi-view triangulation for one-shot Gaussian initialization, followed by differentiable 3DGS optimization and real-time tracking.}%
    \label{fig:domain}%
  \end{center}%
  \vspace{2pt}%
}{}{}
\makeatother

\begin{document}
\maketitle

\begin{abstract}
Gaussian Splatting enables high-fidelity differentiable SLAM by representing scenes as anisotropic Gaussian primitives. Most systems rely on residual-driven densification, which couples approximation refinement with representation growth: accuracy saturates once a moderate number of primitives is reached, yet memory grows throughout the trajectory, causing uncontrolled map expansion and degraded throughput. We introduce TopoGS-SLAM, a stationary formulation that decouples refinement from capacity expansion by constructing a spatially balanced Gaussian representation at initialization and optimizing under fixed topology, turning mapping into a fixed-dimensional problem with bounded size and stable scaling. On standard indoor benchmarks, TopoGS-SLAM matches or surpasses densification-based pipelines in localization and reconstruction accuracy while substantially improving memory scalability and sustained throughput, showing that iterative topology mutation is not required for accurate Gaussian SLAM.
\end{abstract}

\begin{figure}[t]
    \centering
    \includegraphics[width=\linewidth]{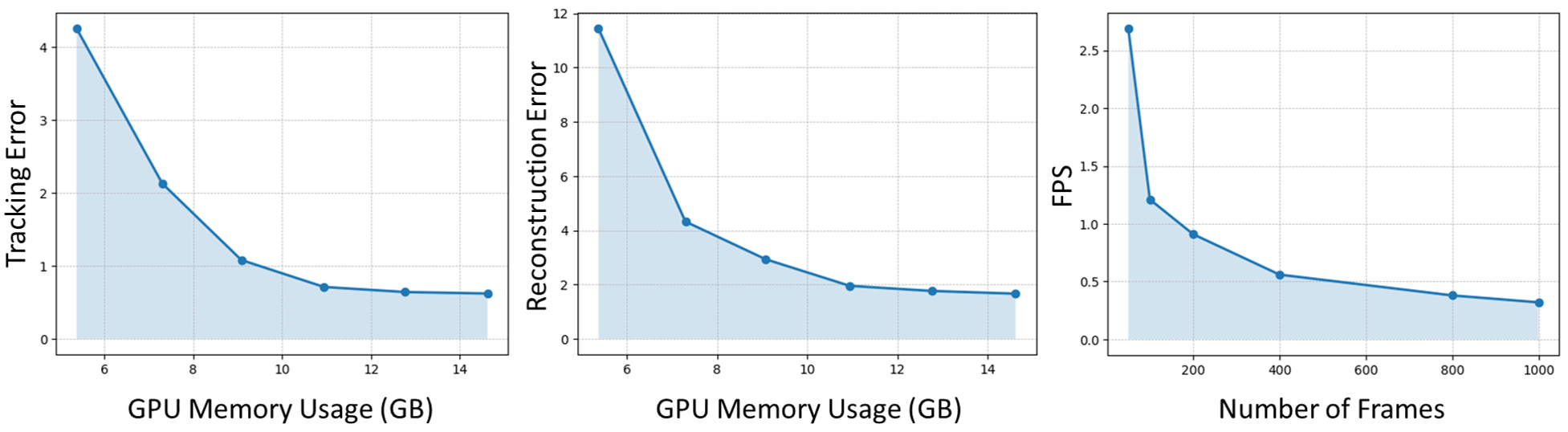}
    \caption{Empirical scaling behavior in densification-based Gaussian SLAM. Tracking and reconstruction errors saturate with increasing memory usage, while throughput degrades as the trajectory grows, indicating uncontrolled capacity expansion.}
    \label{fig:problem}
\end{figure}

\section{Introduction}\label{sec:intro}
Dense SLAM has advanced rapidly through Gaussian Splatting (GS), which represents scenes as anisotropic Gaussians and enables joint tracking and photorealistic reconstruction within a unified differentiable framework. Most GS-SLAM pipelines rely on residual-driven densification that continuously expands Gaussian topology during optimization. As shown in Figure~\ref{fig:problem}, tracking and reconstruction accuracy saturate once a moderate memory footprint is reached, yet the number of primitives and GPU memory grow monotonically with trajectory length while throughput degrades. This reveals a capacity--refinement coupling: topology mutation keeps inflating the representation long after accuracy has stabilized, yielding unstable scaling in long-horizon SLAM.

We ask whether iterative topology expansion is necessary at all. If accurate mapping follows from sufficient spatial coverage, continued mutation is redundant. We propose TopoGS-SLAM, which builds a spatially balanced Gaussian configuration at initialization via Dense Feature Matching (DFM), illustrated in Figure~\ref{fig:domain}, and then optimizes under fixed topology. Decoupling structural expansion from parameter refinement turns mapping into a constant-dimensional problem with bounded growth and stable runtime. Our contributions are:

\begin{itemize}
\item \textbf{Coverage-aware dense initialization.}
A DFM-based multi-view triangulation scheme that builds a spatially balanced Gaussian set in a single pass, removing residual-driven topology expansion.

\item \textbf{Decoupled refinement and capacity.}
We identify the capacity--refinement coupling of densification and recast GS-SLAM as a stationary, fixed-topology optimization problem.

\item \textbf{Stabilized scaling.}
Removing dynamic structural updates yields bounded representation size and sustained throughput over long trajectories.

\item \textbf{High-fidelity reconstruction.}
Fixed-topology initialization preserves geometric consistency and coverage, enabling accurate, complete reconstruction without densification.
\end{itemize}

\section{Related Work}
3D Gaussian Splatting enables real-time view synthesis through anisotropic splats and visibility-aware rendering~\cite{kerbl2023gaussian}, with most implementations relying on residual-driven densification to expand topology during optimization. Several SLAM systems adopt Gaussian maps, including GS-SLAM, MonoGS, SplaTAM, Gauss-SLAM, LoopSplat, and Splat-SLAM~\cite{yan2024gsslam,matsuki2024monogs,keetha2024splatam,zhu2024loopsplat,sandstrom2024splatslam}, typically combining topology growth with loop closure or global pose refinement. Differentiable rendering further enables gradient-based tracking and mapping in Photo-SLAM, GLORIE-SLAM, and Point-SLAM~\cite{huang2024photoslam,zhang2024glorieslam,sandstrom2023pointslam}. Learned feature matching and dense correspondence methods such as SuperPoint, SuperGlue, LoFTR, LightGlue, and DKM~\cite{detone2018superpoint,sarlin2020superglue,sun2021loftr,lindenberger2023lightglue,edstedt2023dkm} improve geometric coverage, while classical Structure-from-Motion and multi-view stereo provide stable geometric priors~\cite{schonberger2016sfm,yao2018mvsnet}. In contrast, our framework constructs a dense Gaussian set upfront from multi-view correspondences and performs mapping under fixed topology without residual-driven densification.

\section{Method}\label{sec:method}
\begin{figure*}[!t]
    \centering
    \includegraphics[width=\textwidth]{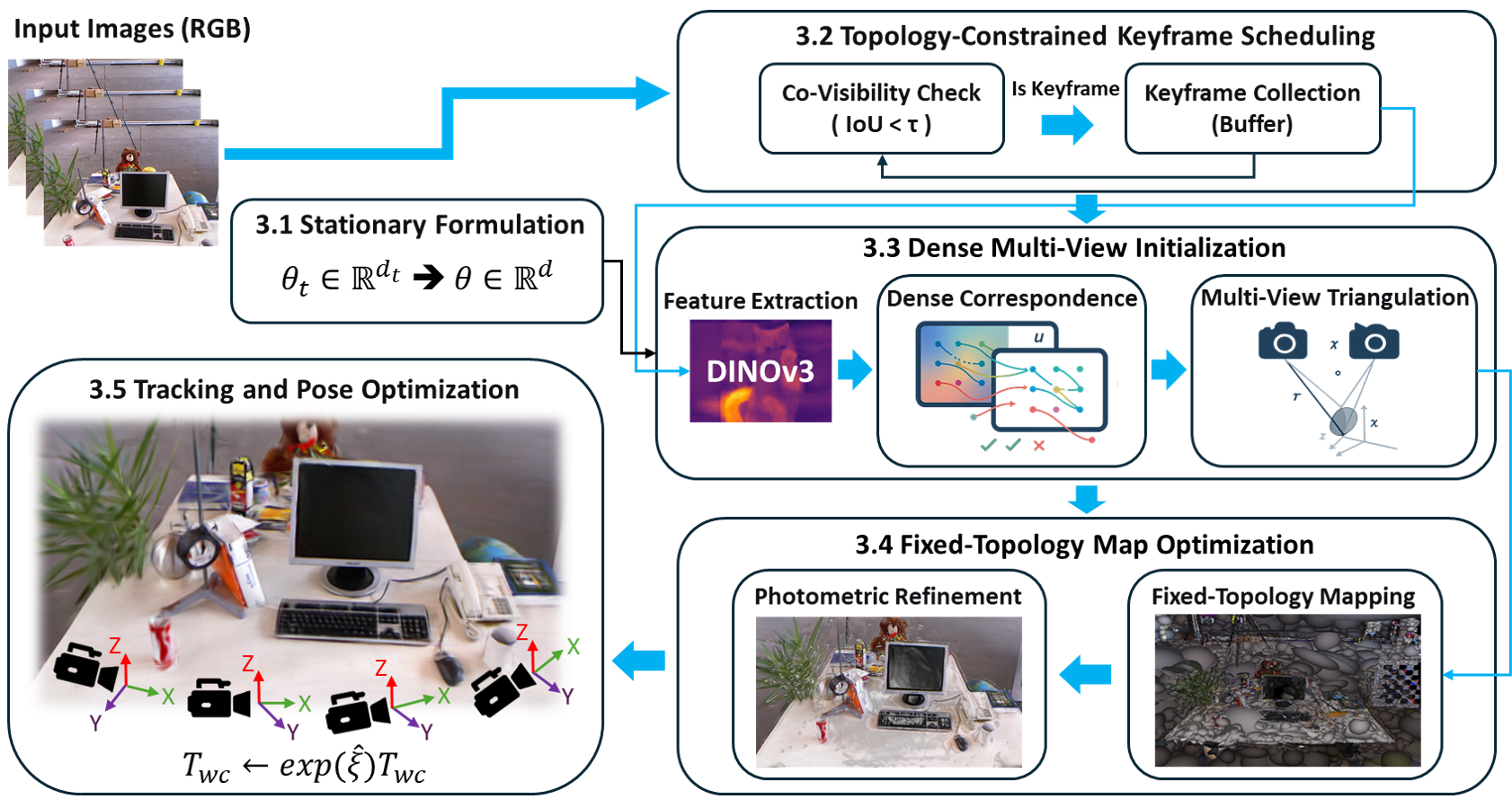}
    \caption{Stationary TopoGS-SLAM pipeline. Coverage-aware keyframe scheduling and one-shot dense initialization define a fixed Gaussian set, followed by photometric mapping and analytic $\mathrm{SE}(3)$ pose refinement without topology expansion.}
    \label{fig:arch}
\end{figure*}

We formulate TopoGS-SLAM as a stationary Gaussian SLAM framework grounded in coverage sufficiency and fixed-topology optimization. Figure~\ref{fig:arch} provides an overview of the complete pipeline.

\subsection{Stationary Reformulation of Gaussian SLAM}
\label{sec:dynamic_stationary}

\paragraph{Dynamic Topology and Non-Stationary Optimization.}

Gaussian SLAM represents a scene as a collection of anisotropic Gaussian primitives
$\mathcal{G}_t=\{G_i\}_{i=1}^{N_t}$
with stacked parameters $\theta_t$ and camera pose $T$.
Rendering is performed via differentiable splatting and alpha compositing, yielding the photometric objective:
\begin{equation}
\mathcal{L}_t(\theta_t,T)
=\big\|\mathcal{S}(\mathcal{G}_t,T)-\bar I\big\|_1.
\end{equation}

In densification-based systems, new primitives are inserted whenever reconstruction residuals exceed a threshold,
\begin{equation}
N_{t+1}=N_t + k_t, \qquad k_t \ge 0,
\end{equation}
which induces an expanding parameter space:
\begin{equation}
\theta_t \in \mathbb{R}^{d_t}, \qquad d_t \propto N_t,
\end{equation}
and therefore a sequence of growing optimization domains:
\begin{equation}
\mathbb{R}^{d_1} \rightarrow \mathbb{R}^{d_2} \rightarrow \cdots.
\end{equation}
Each topology update modifies the Hessian structure, so both dimensionality and curvature evolve during mapping.
The optimization landscape is intrinsically non-stationary.
Representation growth becomes tightly coupled with refinement, and computational cost increases monotonically with trajectory length, even after geometric coverage has effectively saturated.

\paragraph{Fixed Topology as a Stationary Optimization System.}

We instead reformulate Gaussian SLAM under a stationary topology.
The Gaussian set is fixed at initialization,
\begin{equation}
\mathcal{G}=\{G_i\}_{i=1}^{N},
\end{equation}
and only primitive parameters and camera poses are optimized,
\begin{equation}
\theta \in \mathbb{R}^{d}, \qquad d \propto N.
\end{equation}
The global objective over a window $\mathcal{W}$ becomes:
\begin{equation}
\mathcal{L}(\theta)
=\sum_{I\in\mathcal{W}}
\big\|\mathcal{S}(\mathcal{G},T_I)-\bar I_I\big\|_1.
\end{equation}
The parameter space remains constant throughout mapping.
Optimization is therefore performed in a fixed-dimensional nonlinear least-squares system with invariant curvature structure.
Capacity selection is separated from parameter refinement, yielding stationary optimization dynamics and predictable computational scaling.

\paragraph{Representation Capacity and Spatial Coverage.}

A key question is whether a fixed topology provides sufficient representational capacity.
We formalize this through spatial coverage.
Let $\mathcal{M}\subset\mathbb{R}^3$ denote a compact scene surface with finite area $|\mathcal{M}|$, and let $\{\mu_i\}_{i=1}^N$ be the Gaussian centers.
The set forms a $\delta$-cover of $\mathcal{M}$ if:
\begin{equation}
\forall x\in\mathcal{M}, \quad
\exists\, \mu_i \text{ such that } \|x-\mu_i\|\le\delta.
\end{equation}

\paragraph{Capacity Scaling Law.}

If the Gaussian centers form such a $\delta$-cover over a two-dimensional manifold surface, the required number of primitives satisfies the quadratic scaling law:
\begin{equation}
N = \Theta\!\left(\frac{|\mathcal{M}|}{\delta^2}\right),
\end{equation}
which implies that the primitive count scales inversely with the squared resolution:
\begin{equation}
N \propto \frac{1}{\delta^2}.
\end{equation}
Once the desired spatial resolution $\delta$ is specified, the primitive count $N$ is determined and does not depend on trajectory length.

Assume further that the radiance function $f:\mathcal{M}\rightarrow\mathbb{R}^3$ is $L$-Lipschitz continuous,
\begin{equation}
|f(x)-f(y)| \le L\|x-y\|.
\end{equation}
Let $\hat f$ denote the Gaussian kernel approximation with bandwidth $\sigma$ satisfying $\sigma \le c_\sigma \delta$.
Then the approximation error satisfies:
\begin{equation}
\sup_{x\in\mathcal{M}}
|f(x)-\hat f(x)|
\le C_1 L\delta + C_2 \sigma^2,
\end{equation}
where $C_1$ and $C_2$ depend only on the kernel shape.

This bound shows that once spatial coverage is sufficient, reconstruction error becomes independent of further primitive insertion. Additional topology expansion mainly increases computational cost with marginal approximation gain. Under this regime, Gaussian SLAM reduces to refinement in a fixed-dimensional space, whereas residual-driven densification continues to expand dimensionality beyond saturation. The stationary reformulation thus enforces bounded representation capacity while preserving approximation fidelity.

\subsection{Keyframe Scheduling by Co-Visibility}
\label{sec:keyframe}

Under the stationary formulation in Sec.~\ref{sec:dynamic_stationary}, 
representation capacity is determined at initialization rather than through incremental densification. 
Keyframe selection therefore plays a structural role in shaping view diversity and parallax, 
which influence how uniformly the scene surface is observed prior to primitive instantiation.

Let $I_{k^\star}$ denote the most recently accepted keyframe. 
We measure co-visibility between the current frame $I_k$ and $I_{k^\star}$ using the intersection-over-union of visible Gaussians,
\begin{equation}
\mathrm{IoU}(I_k,I_{k^\star}) 
= 
\frac{|V(I_k)\cap V(I_{k^\star})|}
{|V(I_k)\cup V(I_{k^\star})|},
\label{eq:iou}
\end{equation}
where $V(I)$ denotes the set of Gaussians whose screen-space opacity exceeds a small threshold over sufficient image support.

A new keyframe is inserted when $\mathrm{IoU}(I_k,I_{k^\star}) < \tau$ and the inter-view parallax exceeds a predefined bound. 
This condition promotes geometric diversity rather than temporal redundancy, 
so that newly accepted views contribute novel surface observations.

Accepted keyframes are maintained in a bounded buffer $\mathcal{B}$, 
which provides spatially diverse neighboring views for subsequent multi-view initialization.

\subsection{Dense Multi-View Initialization}
\label{sec:init}

To realize stationary Gaussian SLAM, representation capacity must be determined prior to optimization rather than expanded through residual-driven densification. We therefore construct the Gaussian set in a single dense multi-view initialization stage based on DFM. Dense descriptors are extracted using DINOv3~\cite{Simeoni2025dinov3} to establish multi-view correspondences across keyframes. Confidence-aware filtering removes unreliable matches, ensuring geometrically consistent seeds.

\paragraph{Dense Correspondence.}
Let $I_r$ denote the current keyframe and $\mathcal{N}_r \subset \mathcal{B}$ the set of $K$ neighboring keyframes selected by pose proximity and parallax. 
For each pair $(r,n)$ with $n \in \mathcal{N}_r$, a dense matcher predicts a displacement field $\mathbf{u}_{r\to n}(p)$ and confidence map $\kappa_{r\to n}(p)\in[0,1]$. 
Each match is represented as:
\begin{equation}
\big(p,\; p+\mathbf{u}_{r\to n}(p)\big),
\label{eq:corr}
\end{equation}
and filtered via symmetric epipolar consistency and spatial thinning. 
Per-view confidence is aggregated as:
\begin{equation}
\bar\kappa(p)=\frac{1}{|\mathcal{N}_r|}\sum_{n\in\mathcal{N}_r}\kappa_{r\to n}(p).
\label{eq:conf}
\end{equation}

\paragraph{Multi-View Triangulation.}
For each retained correspondence, we perform calibrated linear triangulation. 
Given projection matrices $P_r, P_n \in \mathbb{R}^{3\times4}$ and homogeneous pixels $\tilde x_r,\tilde x_n$, we solve:
\begin{equation}
A \tilde X = 0, \quad 
\tilde X = \arg\min_{\|\tilde X\|=1} \|A\tilde X\|_2,
\end{equation}
where $\tilde X$ is the right singular vector associated with the smallest singular value. 
The Euclidean estimate is $\hat X=\tilde X_{1:3}/\tilde X_4$. 
Among candidate views, we retain the hypothesis with minimal reprojection error and sufficient baseline.

\paragraph{Gaussian Parameter Initialization.}
Each valid triangulation instantiates a Gaussian $G_i$ with:
\begin{equation}
\mu_i^{W}=\hat X(p).
\end{equation}
A local frame $U_i=[\mathbf{t}_1,\mathbf{t}_2,\mathbf{v}]$ is constructed using estimated surface normals, and the covariance is initialized as:
\begin{equation}
\Sigma_i^{W}=U_i\,\mathrm{diag}(s_{\perp}^2,\ s_{\perp}^2,\ s_{\parallel}^2)\,U_i^{\!\top},
\end{equation}
where $s_{\perp}$ reflects image-plane uncertainty and $s_{\parallel}$ accounts for triangulation ambiguity. 
Color is assigned by exposure-compensated sampling, and opacity follows $\bar\kappa(p)$. 
Spatial subsampling enforces balanced coverage before insertion.

This one-shot instantiation fixes the primitive count and spatial allocation prior to photometric refinement, thereby defining representation capacity independently of trajectory length.

\subsection{Fixed-Topology Map Optimization}
\label{sec:mapping}

Each accepted keyframe triggers one-shot dense initialization (Sec.~\ref{sec:init}), after which all primitives participate in refinement under fixed topology. No residual-driven expansion is performed during tracking or mapping.

\paragraph{Redundancy Suppression.}
Each Gaussian $G_i$ maintains observation count $m_i$, cumulative visibility $v_i$, and a moving average of its screen-space footprint. 
To suppress unstable or redundant primitives, periodic cleanup removes splats satisfying:
\begin{equation}
\begin{aligned}
m_i &< m_{\min}, \qquad v_i < v_{\min}, \\
\operatorname{tr}\!\big(\Sigma_i^{W}\big) &> \sigma_{\max}^{2}, \qquad \alpha_i < \alpha_{\min}.
\end{aligned}
\end{equation}
Neighbouring Gaussians with overlapping footprints and similar appearance are merged via visibility-weighted updates. 
These operations only reduce redundancy and do not introduce new primitives, preserving the predefined capacity established during initialization.

\paragraph{Sliding-Window Photometric Optimization.} Let $\mathcal{W}$ denote a window centered at the latest keyframe. 
We jointly refine camera poses $\{T_{WC}\}_{I\in\mathcal{W}}$ and Gaussian parameters $\{c_i,\alpha_i,\mu_i^{W},\Sigma_i^{W}\}$ by minimizing:
\begin{equation}
\mathcal{L}
=
\sum_{I\in\mathcal{W}}
\lambda_{\mathrm{pho}}
\big\| g_I\,\mathcal{S}(\mathcal{G},T_{WC}) + b_I - \bar I \big\|_{1}
+ \mathcal{R},
\label{eq:map-loss}
\end{equation}
where $g_I$ and $b_I$ model exposure variation. 
The regularizer is defined as:
\begin{equation}
\begin{aligned}
\mathcal{R}
&= \lambda_{\mathrm{iso}} \sum_{i}
   \Big\| \Sigma_i^{W} - \tfrac{\mathrm{tr}(\Sigma_i^{W})}{3} I_3 \Big\|_{F} \\
&\quad + \lambda_{\alpha} \sum_{i} \psi(\alpha_i)
   + \lambda_{\mu} \sum_{i} \big\| \mu_i^{W} - \bar\mu_i^{W} \big\|_{2}^{2}
\end{aligned}
\label{eq:reg}
\end{equation}
This term stabilizes anisotropy, opacity, and early updates. 
Optimization alternates between pose refinement and joint parameter updates using analytic splatting and pose Jacobians.

Since topology remains fixed, refinement operates in a constant-dimensional space, and computational complexity scales with the predefined primitive count rather than trajectory length.

\subsection{Fixed-Topology Tracking \& Pose Optimization}
\label{sec:track}

For each incoming frame, the camera pose is estimated by minimizing a photometric objective defined through the differentiable 3DGS renderer. Let $I(\mathcal{G},T_{WC})=\mathcal{S}(\mathcal{G},T_{WC})$ denote the rendered image and $\bar I$ the observation. We minimize:
\begin{equation}
E_{\text{pho}}=\big\| I(\mathcal{G},T_{WC})-\bar I \big\|_{1},
\label{eq:photometric}
\end{equation}
augmented with an affine brightness model $gI(\cdot)+b$ to account for exposure variation. Pixels with low opacity or weak gradients are down-weighted. A fixed number of gradient updates is performed per frame.

Rendering follows standard alpha compositing in screen space, where projected Gaussians are rasterized in front-to-back order and occlusion is handled implicitly through transmittance. No explicit depth map is required.

Camera poses are updated on $\mathrm{SE}(3)$ using a left-multiplicative twist:
\begin{equation}
T_{WC} \leftarrow \exp(\hat\xi)\,T_{WC},
\end{equation}
with optimization in minimal coordinates. Let $\mu^{C}=R\mu^{W}+t$ denote the camera-space mean of a Gaussian. The analytic Jacobian with respect to the pose twist is:
\begin{equation}
\frac{\partial \mu^{C}}{\partial \xi}
=
\big[ I \ \ -[\mu^{C}]_{\times} \big],
\end{equation}
and projection yields:
\begin{equation}
\frac{\partial \mu^{I}}{\partial \xi}
=
J_{\pi}(\mu^{C})
\big[ I \ \ -[\mu^{C}]_{\times} \big].
\end{equation}
Gradients propagate through splatting and projection without generic autodiff, enabling efficient and stable tracking.

Since topology remains fixed, pose refinement operates in a constant-dimensional parameter space, and per-frame computational cost scales with the predefined primitive count.

\section{Experiments}\label{sec:experiment}
\subsection{Experimental Setup}\label{sec:exp_setup}

\paragraph{Datasets.} We evaluate on TUM RGB-D and Replica. TUM RGB-D is tested in both monocular and RGB-D modes. Replica is used for photometric evaluation on \emph{room0–2} and \emph{office0–4}, following standard splits for comparable rendering and runtime analysis.

\paragraph{Implementation.} Gaussian rasterization and analytic gradients are implemented in CUDA, with the remaining pipeline in PyTorch. Mixed precision is enabled where appropriate. Tracking runs in real time, while mapping executes asynchronously within a bounded local window. Non-standard hyperparameters are provided in the supplementary.

\paragraph{Evaluation Metrics.} Tracking is evaluated using RMSE of Absolute Trajectory Error (ATE) on keyframes. Photometric quality is measured by PSNR~\cite{hore2010psnr}, SSIM~\cite{wang2004ssim}, and LPIPS~\cite{zhang2018lpips}. Reconstruction reports \emph{Acc.} [cm]$\downarrow$, \emph{Comp.} [cm]$\downarrow$, and \emph{Comp.Ratio} (\%)$\uparrow$. Unless stated otherwise, we sample 50K surface points with $\tau=5$\,cm and report per-scene averages. Photometric metrics are computed every fifth non-keyframe. All experiments are repeated three times and averaged.

\subsection{Optimization Time and Convergence Analysis}\label{sec:exp_model_training}
Our method replaces residual-driven densification with DFM-based single-step initialization while preserving the original optimization configuration. All optimizer settings, schedules, and iteration budgets remain identical, modifying only the topology refinement mechanism. Experiments are conducted on an NVIDIA RTX 3090 GPU under the same stopping criteria. On TUM RGB-D (Table~\ref{tab:tum_train}), the average optimization time decreases from 14.8 to 6.3 minutes without altering the objective or rendering formulation. Because the Gaussian set is fixed at initialization, the optimizer operates on a constant parameter dimension throughout the sequence, so the photometric loss decreases smoothly without the periodic perturbations that densification events introduce into the optimization landscape.

\begin{figure*}[t]
  \centering
  \includegraphics[width=\textwidth]{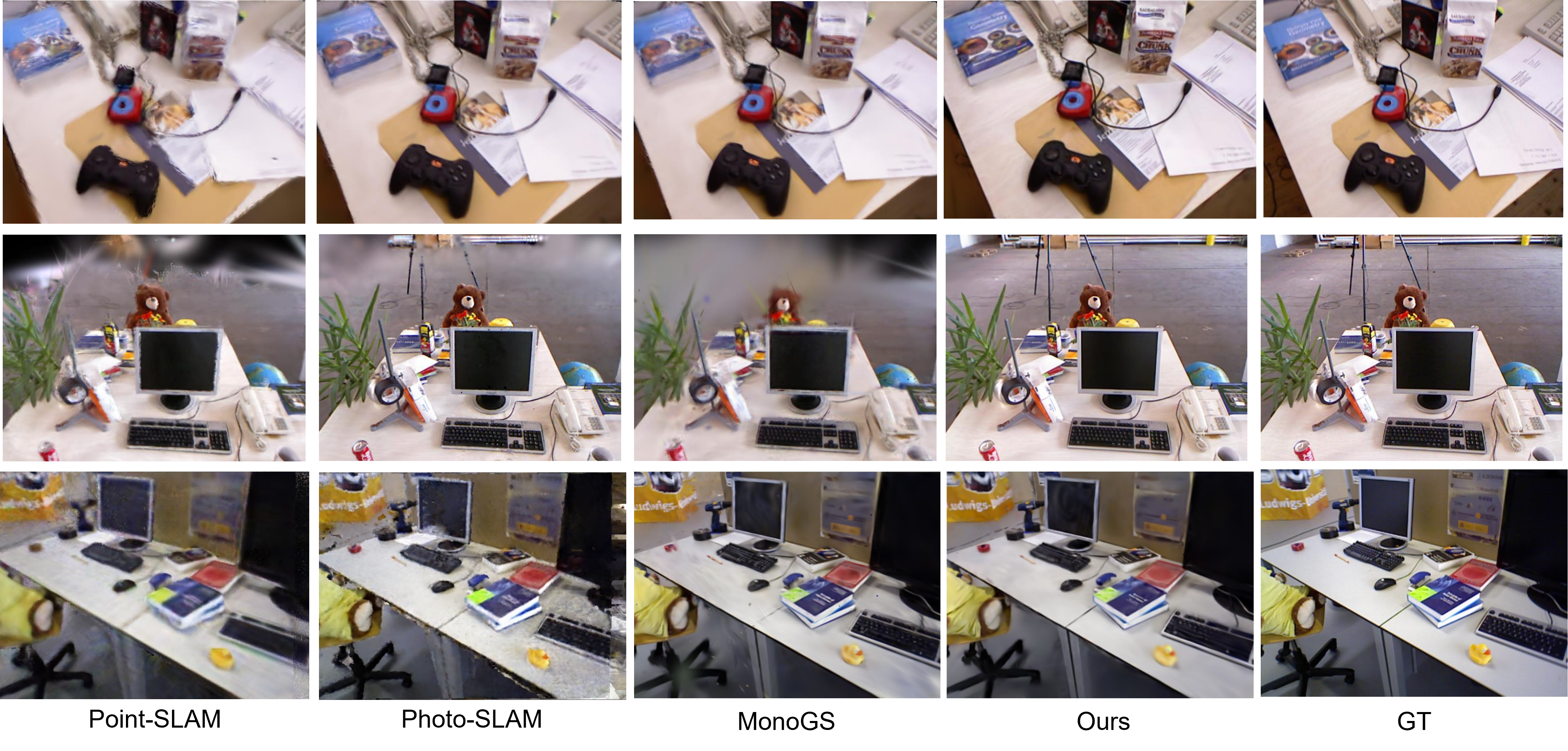}
  \caption{Rendering results on the TUM dataset. Keyframe-triggered initialization produces sharper edges and fewer artifacts than residual-driven densification.}
  \label{fig:qual}
\end{figure*}

\begin{figure*}[t]
  \centering
  \includegraphics[width=\textwidth]{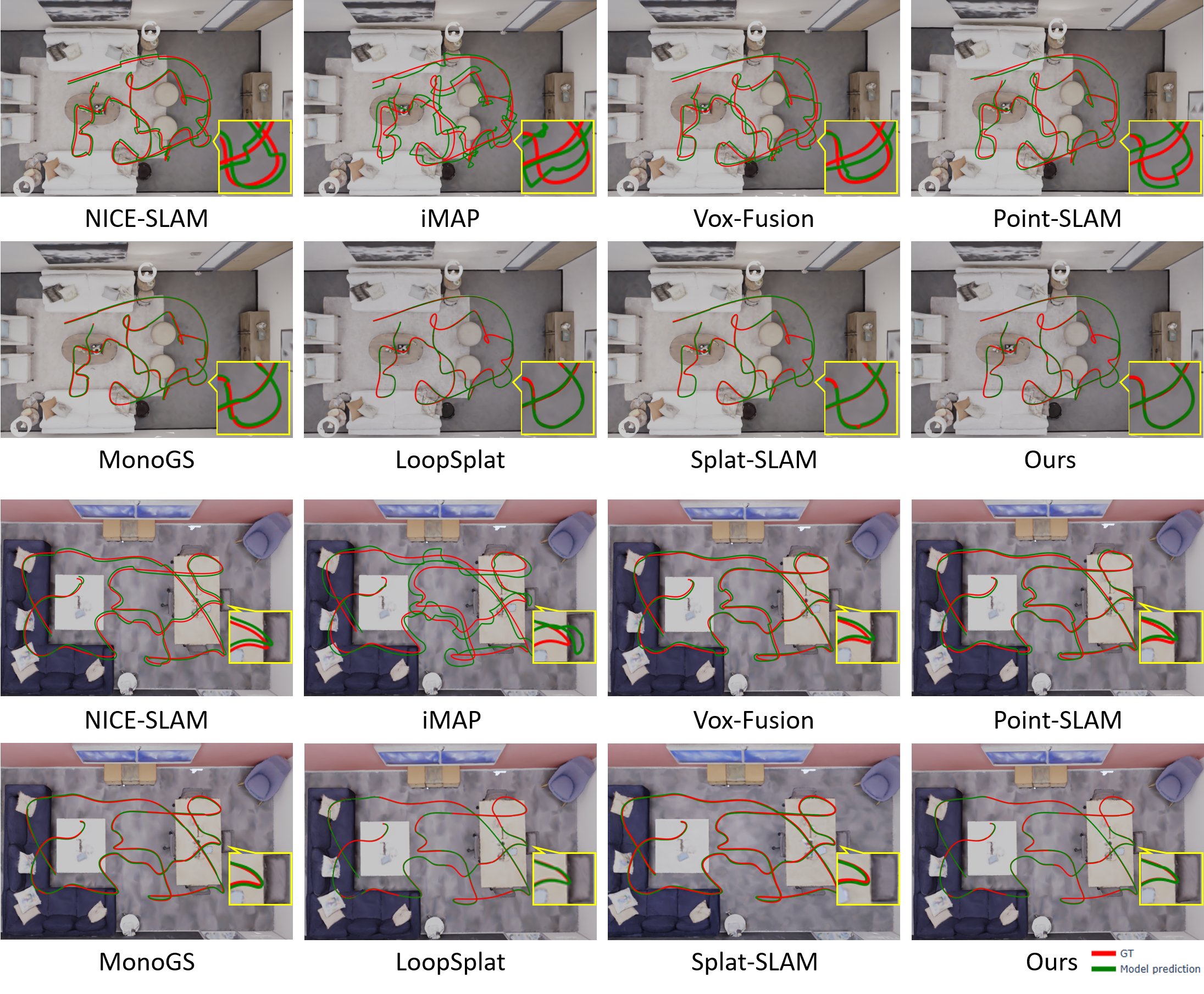}
  \caption{Trajectory comparison on Replica \textit{room0}. The red line indicates the ground-truth path and the green line shows the estimated trajectory. Our method aligns more closely with the ground-truth and exhibits fewer large drifts than previous systems.}
  \label{fig:traj_compare}
\end{figure*}

\begin{table}[b]
\centering
\caption{Optimization time (minutes) on TUM RGB-D \textit{fr1/desk}, \textit{fr2/xyz}, and \textit{fr3/office}, reported per sequence and on average; lower is faster.}
\label{tab:tum_train}
\centering
\setlength{\tabcolsep}{6pt}
\renewcommand{\arraystretch}{1.10}
\begin{adjustbox}{max width=\linewidth}
\begin{tabular}{lcccc}
\toprule
Method & \textit{fr1/desk} & \textit{fr2/xyz} & \textit{fr3/office} & Avg. \\
\midrule
MonoGS~\cite{matsuki2024monogs} & 6.4 & 20.6 & 17.5 & 14.8 \\
LoopSplat~\cite{zhu2024loopsplat} & 5.9 & 19.4 & 18.1 & 14.5 \\
Splat-SLAM~\cite{sandstrom2024splatslam} & 5.7 & 18.6 & 17.4 & 13.9 \\
Ours (RGB) & \textbf{2.9} & \textbf{8.0} & \textbf{8.4} & \textbf{6.3} \\
\bottomrule
\end{tabular}
\end{adjustbox}
\end{table}

\subsection{Rendering Quality and Throughput}\label{sec:exp_loc_render}
\begin{table}[b]
\caption{Rendering quality on TUM RGB-D (monocular) over \textit{fr1/desk}, \textit{fr2/xyz}, and \textit{fr3/office}, reported per sequence and averaged.}
\label{tab:tum_render}
\centering
\scriptsize
\setlength{\tabcolsep}{4.7pt}
\renewcommand{\arraystretch}{0.88}
\setlength{\aboverulesep}{0pt}
\setlength{\belowrulesep}{0pt}
\setlength{\belowbottomsep}{0pt}
\begin{adjustbox}{max width=\linewidth}
\begin{tabular}{l l c c c c}
\toprule
Method & Metric & \textit{fr1/desk} & \textit{fr2/xyz} & \textit{fr3/office} & Avg. \\
\midrule
\multirow{3}{*}{Point\mbox{-}SLAM~\cite{sandstrom2023pointslam}}
& PSNR[dB]$\uparrow$ & 13.79 & 17.62 & 18.29 & 16.57 \\
& SSIM$\uparrow$     & 0.625 & 0.710 & 0.749 & 0.695 \\
& LPIPS$\downarrow$  & 0.545 & 0.584 & 0.452 & 0.527 \\
\midrule
\multirow{3}{*}{Photo\mbox{-}SLAM~\cite{huang2024photoslam}}
& PSNR[dB]$\uparrow$ & 20.97 & 21.07 & 19.59 & 20.54 \\
& SSIM$\uparrow$     & 0.740 & 0.730 & 0.690 & 0.720 \\
& LPIPS$\downarrow$  & 0.230 & 0.170 & 0.240 & 0.213 \\
\midrule
\multirow{3}{*}{MonoGS~\cite{matsuki2024monogs}}
& PSNR[dB]$\uparrow$ & 19.67 & 16.17 & 20.63 & 18.82 \\
& SSIM$\uparrow$     & 0.730 & 0.720 & 0.770 & 0.740 \\
& LPIPS$\downarrow$  & 0.330 & 0.310 & 0.340 & 0.327 \\
\midrule
\multirow{3}{*}{SplaTAM~\cite{keetha2024splatam}}
& PSNR[dB]$\uparrow$ & 21.49 & 25.06 & 21.17 & 22.57 \\
& SSIM$\uparrow$     & 0.839 & \textbf{0.950} & 0.861 & \textbf{0.883} \\
& LPIPS$\downarrow$  & 0.255 & \textbf{0.099} & 0.221 & \textbf{0.192} \\
\midrule
\multirow{3}{*}{RK\mbox{-}SLAM~\cite{ma2025rkslam}}
& PSNR[dB]$\uparrow$ & 22.31 & 22.47 & 20.67 & 21.82 \\
& SSIM$\uparrow$     & 0.741 & 0.729 & 0.710 & 0.727 \\
& LPIPS$\downarrow$  & 0.254 & 0.220 & 0.251 & 0.242 \\
\midrule
\multirow{3}{*}{LoopSplat~\cite{zhu2024loopsplat}}
& PSNR[dB]$\uparrow$ & 22.03 & 22.68 & 23.47 & 22.72 \\
& SSIM$\uparrow$     & 0.849 & 0.892 & \textbf{0.879} & 0.873 \\
& LPIPS$\downarrow$  & 0.307 & 0.217 & 0.253 & 0.259 \\
\midrule
\multirow{3}{*}{Splat\mbox{-}SLAM~\cite{sandstrom2024splatslam}}
& PSNR[dB]$\uparrow$ & 23.04 & \textbf{25.42} & \textbf{25.02} & \textbf{24.49} \\
& SSIM$\uparrow$     & 0.782 & 0.796 & 0.826 & 0.801 \\
& LPIPS$\downarrow$  & \textbf{0.182} & 0.235 & \textbf{0.199} & 0.205 \\
\midrule
\multirow{3}{*}{Ours (RGB)}
& PSNR[dB]$\uparrow$ & \textbf{23.11} & 24.85 & 23.59 & 23.85 \\
& SSIM$\uparrow$     & \textbf{0.853} & 0.896 & 0.801 & 0.850 \\
& LPIPS$\downarrow$  & 0.232 & 0.198 & 0.219 & 0.216 \\
\bottomrule
\end{tabular}
\end{adjustbox}
\end{table}

\begin{table}[b]
\centering
\caption{Rendering quality on Replica (RGB-D and monocular) over \textit{room0--2} and \textit{office0--4}. PSNR$\uparrow$, SSIM$\uparrow$, and LPIPS$\downarrow$ are averaged over all sequences; full-pipeline FPS is shown in parentheses when available.}
\label{tab:replica_render}
\centering
\scriptsize
\setlength{\tabcolsep}{2.3pt}
\renewcommand{\arraystretch}{1.22}
\setlength{\aboverulesep}{0pt}
\setlength{\belowrulesep}{0pt}
\setlength{\belowbottomsep}{0pt}

\begin{adjustbox}{max width=\linewidth}
\begin{tabular}{l l c c c c c c c c c}
\toprule
Method (FPS) & Metric & room0 & room1 & room2 & office0 & office1 & office2 & office3 & office4 & Avg. \\
\midrule
\multirow{3}{*}{\begin{tabular}[t]{@{}l@{}}NICE\mbox{-}SLAM~\cite{zhu2022niceslam}\end{tabular}}
& PSNR[dB]$\uparrow$ & 22.12 & 22.47 & 24.52 & 29.07 & 30.34 & 19.66 & 22.23 & 24.94 & 24.42 \\
& SSIM$\uparrow$     & 0.689 & 0.757 & 0.814 & 0.874 & 0.868 & 0.797 & 0.801 & 0.856 & 0.809 \\
& LPIPS$\downarrow$  & 0.330 & 0.271 & 0.208 & 0.229 & 0.181 & 0.235 & 0.209 & 0.198 & 0.233 \\
\midrule
\multirow{3}{*}{\begin{tabular}[t]{@{}l@{}}Vox\mbox{-}Fusion~\cite{yang2022voxfusion}\end{tabular}}
& PSNR[dB]$\uparrow$ & 22.39 & 22.36 & 23.92 & 27.79 & 29.83 & 20.33 & 23.47 & 25.21 & 24.41 \\
& SSIM$\uparrow$     & 0.683 & 0.751 & 0.798 & 0.857 & 0.876 & 0.794 & 0.803 & 0.847 & 0.801 \\
& LPIPS$\downarrow$  & 0.303 & 0.269 & 0.234 & 0.241 & 0.184 & 0.243 & 0.213 & 0.199 & 0.236 \\
\midrule
\multirow{3}{*}{\begin{tabular}[t]{@{}l@{}}Point\mbox{-}SLAM~\cite{sandstrom2023pointslam}\end{tabular}}
& PSNR[dB]$\uparrow$ & 32.40 & 34.08 & 35.50 & 38.26 & 39.16 & 33.98 & 33.48 & 33.49 & 35.17 \\
& SSIM$\uparrow$     & 0.974 & \textbf{0.977} & 0.979 & 0.982 & 0.986 & 0.962 & 0.960 & 0.979 & 0.975 \\
& LPIPS$\downarrow$  & 0.113 & 0.116 & 0.111 & 0.100 & 0.118 & 0.156 & 0.132 & 0.142 & 0.124 \\
\midrule
\multirow{3}{*}{Co\mbox{-}SLAM~\cite{wang2023coslam}}
& PSNR[dB]$\uparrow$ & 28.88 & 28.51 & 29.37 & 35.44 & 34.63 & 26.56 & 28.79 & 32.16 & 28.42 \\
& SSIM$\uparrow$     & 0.892 & 0.843 & 0.851 & 0.854 & 0.826 & 0.814 & 0.866 & 0.856 & 0.837 \\
& LPIPS$\downarrow$  & 0.213 & 0.205 & 0.215 & 0.177 & 0.161 & 0.172 & 0.163 & 0.176 & 0.185 \\
\midrule
\multirow{3}{*}{SplaTAM~\cite{keetha2024splatam}}
& PSNR[dB]$\uparrow$ & 32.49 & 33.72 & 34.65 & 38.29 & 39.04 & 31.91 & 30.05 & 31.83 & 30.98 \\
& SSIM$\uparrow$     & \textbf{0.975} & 0.970 & 0.980 & 0.982 & 0.982 & 0.965 & 0.952 & 0.949 & 0.953 \\
& LPIPS$\downarrow$  & 0.072 & 0.096 & 0.078 & 0.086 & 0.093 & 0.100 & 0.110 & 0.150 & 0.179 \\
\midrule
\multirow{3}{*}{Gauss\mbox{-}SLAM~\cite{yan2024gsslam}}
& PSNR[dB]$\uparrow$ & 29.57 & 31.61 & 33.46 & 38.39 & 39.62 & 32.91 & 33.62 & 34.26 & 30.90 \\
& SSIM$\uparrow$     & 0.944 & 0.952 & 0.973 & 0.985 & \textbf{0.991} & 0.974 & 0.982 & 0.979 & 0.972 \\
& LPIPS$\downarrow$  & 0.197 & 0.184 & 0.148 & 0.099 & 0.097 & 0.158 & 0.123 & 0.138 & 0.229 \\
\midrule
\multirow{3}{*}{\begin{tabular}[t]{@{}l@{}}MonoGS~\cite{matsuki2024monogs}\\(1.04)\end{tabular}}
& PSNR[dB]$\uparrow$ & 34.83 & \textbf{36.43} & \textbf{37.49} & 39.95 & \textbf{42.09} & \textbf{36.24} & \textbf{36.70} & 36.07 & \textbf{37.50} \\
& SSIM$\uparrow$     & 0.954 & 0.959 & 0.965 & 0.971 & 0.974 & 0.964 & 0.963 & 0.957 & 0.960 \\
& LPIPS$\downarrow$  & \textbf{0.068} & \textbf{0.076} & \textbf{0.075} & \textbf{0.072} & \textbf{0.055} & \textbf{0.078} & \textbf{0.065} & 0.099 & \textbf{0.070} \\
\midrule
\multirow{3}{*}{\begin{tabular}[t]{@{}l@{}}LoopSplat~\cite{zhu2024loopsplat}\\(1.10)\end{tabular}}
& PSNR[dB]$\uparrow$ & 32.79 & 35.01 & 35.84 & \textbf{40.45} & 39.92 & 34.31 & 35.28 & \textbf{36.73} & 36.29 \\
& SSIM$\uparrow$     & 0.970 & 0.975 & 0.982 & \textbf{0.989} & 0.987 & \textbf{0.982} & \textbf{0.987} & \textbf{0.986} & \textbf{0.982} \\
& LPIPS$\downarrow$  & 0.122 & 0.129 & 0.118 & 0.092 & 0.130 & 0.148 & 0.103 & 0.113 & 0.119 \\
\midrule
\multirow{3}{*}{\begin{tabular}[t]{@{}l@{}}Splat\mbox{-}SLAM ~\cite{sandstrom2024splatslam}\\(1.24)\end{tabular}}
& PSNR[dB]$\uparrow$ & 32.18 & 34.35 & 35.88 & 38.68 & 39.87 & 35.20 & 35.24 & 35.06 & 35.81 \\
& SSIM$\uparrow$     & 0.905 & 0.929 & 0.943 & 0.955 & 0.969 & 0.962 & 0.952 & 0.975 & 0.949 \\
& LPIPS$\downarrow$  & 0.155 & 0.114 & 0.097 & 0.095 & 0.084 & 0.088 & 0.079 & \textbf{0.057} & 0.084 \\
\midrule
\multirow{3}{*}{\begin{tabular}[t]{@{}l@{}}Ours (RGB)\\(12.05)\end{tabular}}
& PSNR[dB]$\uparrow$ & \textbf{35.95} & 33.55 & 32.45 & 34.45 & 35.45 & 34.87 & 34.02 & 35.85 & 34.57 \\
& SSIM$\uparrow$     & 0.852 & 0.945 & \textbf{0.985} & 0.952 & 0.925 & 0.952 & 0.855 & 0.961 & 0.928 \\
& LPIPS$\downarrow$  & 0.085 & 0.092 & 0.112 & 0.088 & 0.096 & \textbf{0.078} & 0.101 & 0.096 & 0.093 \\
\bottomrule
\end{tabular}
\end{adjustbox}
\end{table}

We report rendering fidelity and full-pipeline throughput in Table~\ref{tab:tum_render} and Table~\ref{tab:replica_render}. On Replica, our method achieves the highest full-pipeline FPS among Gaussian-based SLAM systems, reaching \textbf{12.05 FPS}. Despite this acceleration, it maintains competitive rendering quality with an average PSNR of \textbf{34.57} across \textit{room0--2} and \textit{office0--4}. 
The speed gain arises from eliminating residual-driven densification via single-step initialization, reducing per-frame overhead while preserving the optimization objective and renderer.
On TUM RGB-D, the system runs in real time and achieves PSNR, SSIM, and LPIPS competitive with Gaussian SLAM variants, ranking second in average PSNR. Qualitative comparisons in Figure~\ref{fig:qual} corroborate these numbers: keyframe-triggered initialization yields sharper edges and fewer artifacts.

\subsection{Localization Accuracy}\label{sec:exp_loc}
On Replica, the proposed method achieves an average ATE of \textbf{0.47\,cm} across \textit{room0--2} and \textit{office0--4}, as summarized in Table~\ref{tab:replica_loc}. Low error is maintained consistently across all scenes, including layout variations and loop revisits. Compared with prior GS-based SLAM systems, lower average RMSE is observed while preserving stable performance in each individual sequence. Because the map topology stays fixed throughout tracking, pose optimization operates over a stable set of primitives and avoids the drift that often accompanies on-the-fly densification. This consistency holds even in larger office layouts, where accumulated error typically grows fastest and frame-to-frame tracking becomes most challenging. The bounded error throughout long trajectories indicates robust and drift-resistant pose estimation.

\begin{table}[t]
\caption{Camera tracking results on the Replica dataset under the RGB-D and monocular settings. Reported values denote RMSE of ATE across \emph{room0–2} and \emph{office0–4}.}
\label{tab:replica_loc}
\centering
\scriptsize
\setlength{\tabcolsep}{4.7pt}
\renewcommand{\arraystretch}{1.05}
\begin{adjustbox}{max width=\linewidth}
\begin{tabular}{lccccccccc}
\toprule
Method & room0 & room1 & room2 & office0 & office1 & office2 & office3 & office4 & Avg. \\
\midrule
iMAP~\cite{sucar2021imap} & 3.12 & 2.54 & 2.31 & 1.69 & 1.03 & 3.99 & 4.05 & 1.93 & 2.58 \\
NICE\mbox{-}SLAM~\cite{zhu2022niceslam} & 0.97 & 1.31 & 1.07 & 0.88 & 1.00 & 1.06 & 1.10 & 1.13 & 1.07 \\
Vox\mbox{-}Fusion~\cite{yang2022voxfusion} & 1.37 & 4.70 & 1.47 & 8.48 & 2.04 & 2.58 & 1.11 & 2.94 & 3.09 \\
ESLAM~\cite{johari2023eslam} & 0.71 & 0.70 & 0.52 & 0.57 & 0.55 & 0.58 & 0.72 & 0.63 & 0.63 \\
Point\mbox{-}SLAM~\cite{sandstrom2023pointslam} & 0.61 & 0.41 & \textbf{0.37} & \textbf{0.38} & 0.48 & 0.54 & 0.69 & 0.72 & 0.53 \\
MonoGS~\cite{matsuki2024monogs} & 0.62 & 0.62 & 0.77 & 0.44 & 0.52 & \textbf{0.23} & 0.62 & 2.25 & 0.76 \\
LoopSplat~\cite{zhu2024loopsplat} & 0.58 & 0.83 & 0.62 & 0.60 & 0.56 & 0.53 & 0.52 & \textbf{0.56} & 0.60 \\
Splat\mbox{-}SLAM~\cite{sandstrom2024splatslam} & 0.65 & 0.66 & 0.60 & 0.65 & 0.63 & 0.79 & 0.71 & 0.72 & 0.68 \\
Ours (RGB) & \textbf{0.45} & \textbf{0.40} & 0.53 & 0.52 & \textbf{0.44} & 0.36 & \textbf{0.45} & 0.63 & \textbf{0.47} \\
\bottomrule
\end{tabular}
\end{adjustbox}
\end{table}

\subsection{Reconstruction Fidelity}\label{sec:exp_reconstruction}
We achieve \textbf{1.537\,cm} accuracy, \textbf{1.477\,cm} completeness error, and \textbf{97.843\%} completeness ratio (Table~\ref{tab:recon_metrics}), the best overall performance among compared methods. Our approach outperforms LoopSplat and Splat-SLAM in geometric precision and surface coverage, producing clearer structural delineation and more consistent reconstruction in cluttered regions, as illustrated in the Appendix. These gains are attributed to visibility-aware anisotropic Gaussians and the stationary fixed-topology formulation.

\begin{table}[t]
\caption{Reconstruction results on the Replica dataset. Lower is better for Acc./Comp., higher for Comp.Ratio.}
\label{tab:recon_metrics}
\centering

\footnotesize
\setlength{\tabcolsep}{5pt}
\renewcommand{\arraystretch}{0.95}
\setlength{\aboverulesep}{0pt}
\setlength{\belowrulesep}{0pt}
\setlength{\belowbottomsep}{0pt}

\begin{tabular}{l c c c}
\toprule
\multirow{2}{*}{Methods} & \multicolumn{3}{c}{Reconstruction} \\
\cmidrule(lr){2-4}
 & Acc.\,[cm] $\downarrow$ & Comp.\,[cm] $\downarrow$ & Comp.Ratio (\%) $\uparrow$ \\
\midrule
iMAP~\cite{sucar2021imap}          & 3.624 & 4.934 & 80.515 \\
NICE-SLAM~\cite{zhu2022niceslam}   & 2.373 & 2.645 & 91.137 \\
Vox-Fusion~\cite{yang2022voxfusion}& 1.882 & 2.563 & 90.936 \\
Co-SLAM~\cite{wang2023coslam}      & 2.104 & 2.082 & 93.435 \\
ESLAM~\cite{johari2023eslam}       & 2.082 & 1.754 & 96.427 \\
SNI-SLAM~\cite{zhu2024snislam}     & 1.942 & 1.702 & 96.624 \\
LoopSplat~\cite{zhu2024loopsplat}  & 1.821 & 1.733 & 95.246 \\
Splat-SLAM~\cite{sandstrom2024splatslam}   & 2.436 & 3.644 & 84.692 \\
Ours                               & \textbf{1.537} & \textbf{1.477} & \textbf{97.843} \\
\bottomrule
\end{tabular}

\end{table}

\subsection{Ablation Study}\label{sec:ablation_keyframe_budget}
\paragraph{Effect of Dense Initialization.} Table~\ref{tab:abl_wo} evaluates the effect of removing DFM-based dense initialization. Across all TUM RGB-D sequences, dense initialization substantially improves both rendering quality and geometric accuracy. On \textit{fr1.desk}, PSNR increases from 19.67 to 23.11 and RMSE decreases from 1.50 to 1.02. On \textit{fr2.xyz}, PSNR improves from 16.17 to 24.85 with RMSE reduced from 1.44 to 0.98, with similar trends on \textit{fr3.office}. These results show that multi-view triangulated seeds provide stable early coverage, while residual-driven densification leads to delayed coverage and higher reconstruction error.

\begin{table}[htbp]
\centering
\caption{Impact of DFM on the TUM RGB-D dataset.}
\label{tab:abl_wo}
\footnotesize
\setlength{\tabcolsep}{6pt}
\renewcommand{\arraystretch}{0.95}
\setlength{\aboverulesep}{0pt}
\setlength{\belowrulesep}{0pt}
\setlength{\belowbottomsep}{0pt}
\begin{tabular}{llcccc}
\toprule
& Method & PSNR $\uparrow$ & SSIM $\uparrow$ & LPIPS $\downarrow$ & RMSE $\downarrow$ \\
\midrule
\multirow{2}{*}{\textit{fr1/desk}}
& w/o DFM & 19.67 & 0.73  & 0.33  & 1.50 \\
& Ours    & \textbf{23.11} & \textbf{0.853} & \textbf{0.232} & \textbf{1.02} \\
\midrule
\multirow{2}{*}{\textit{fr2/xyz}}
& w/o DFM & 16.17 & 0.72  & 0.31  & 1.44 \\
& Ours    & \textbf{24.85} & \textbf{0.896} & \textbf{0.198} & \textbf{0.98} \\
\midrule
\multirow{2}{*}{\textit{fr3/office}}
& w/o DFM & 20.63 & 0.77  & 0.34  & 1.49 \\
& Ours    & \textbf{23.59} & \textbf{0.801} & \textbf{0.219} & \textbf{1.05} \\
\bottomrule
\end{tabular}

\vspace{9pt}
\centering
\includegraphics[width=\linewidth]{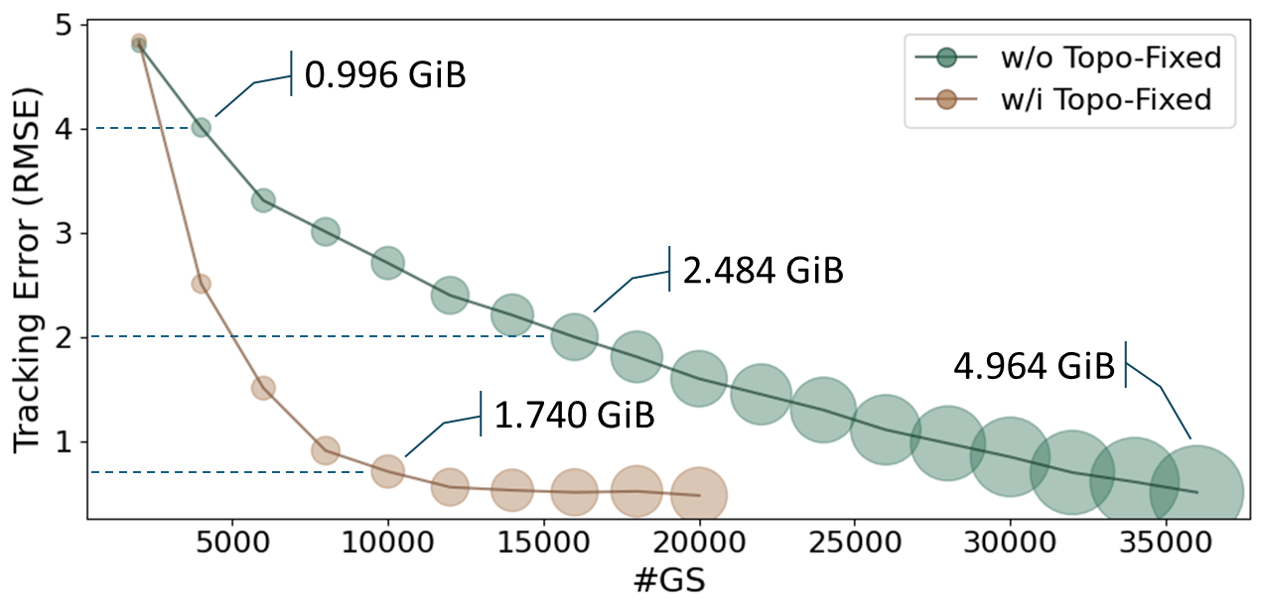}
\captionof{figure}{Effect of topology fixing on tracking and memory scaling. Tracking error vs.\ number of Gaussian primitives; circle radius denotes GPU usage (GiB). Fixed topology attains lower error with fewer primitives.}
\label{fig:ablation_topo}
\end{table}

The effect of topology fixing on tracking accuracy and memory usage is shown in Figure~\ref{fig:ablation_topo}. Tracking error is plotted against the number of Gaussian primitives, with circle radius indicating GPU usage in GiB. With fixed topology, error converges around 10k Gaussians using modest memory. Without topology fixing, error continues decreasing beyond 35k Gaussians without clear saturation, accompanied by steadily increasing memory usage. These results support the coverage-based stationary formulation and indicate that continual topology expansion is unnecessary for accurate tracking.

\subsection{Bounded Capacity and Efficiency Scaling}
\label{sec:efficiency}
We evaluate the joint efficiency and accuracy of TopoGS-SLAM against recent GS-based SLAM systems on Replica (Table~\ref{tab:efficiency}). Under a unified protocol, our method attains the lowest tracking error (0.45 cm) and reconstruction error (1.54 cm) while reducing GPU memory to 5.5\,GiB, map size to 0.64\,MB, and the Gaussian count to 0.01\,M---nearly an order of magnitude below densification-based methods---and sustains the highest throughput at 12.05 FPS. Because topology is fixed, representation size and computational cost remain bounded as the trajectory grows, demonstrating that accurate, scalable Gaussian SLAM is achievable through controlled capacity design rather than continual topology mutation.

\begin{table}[htbp]
\centering
\caption{Efficiency and accuracy comparison on Replica \textit{room0}. All methods are evaluated on an NVIDIA RTX 3090 GPU using single-thread execution for fairness. Lower is better for tracking and reconstruction error, GPU usage, and map size, while higher is better for FPS.}
\label{tab:efficiency}
\setlength{\tabcolsep}{6pt}
\begin{adjustbox}{max width=\linewidth}
\begin{tabular}{lcccccc}
\toprule
Method & Tracking & Recon. & GPU & Map Size & Avg. & \#GS \\
       & Error\,[cm] & Error\,[cm] & (GiB) & (MB) & FPS & (M) \\
\midrule
MonoGS      & 0.62 & 2.07 & 14.62 & 6.8 & 0.32 & 0.11 \\
LoopSplat   & 0.58 & 1.82 & 16.03 & 7.1 & 1.10 & 0.13 \\
Splat-SLAM  & 0.65 & 2.43 & 17.57 & 6.5 & 1.24 & 0.10 \\
\textbf{TopoGS-SLAM} & \textbf{0.45} & \textbf{1.54} & \textbf{5.5} & \textbf{0.64} & \textbf{12.05} & \textbf{0.01} \\
\bottomrule
\end{tabular}
\end{adjustbox}
\end{table}

\section{Conclusion}\label{sec:conclusion}
We revisit the assumption that representation growth must accompany refinement in Gaussian Splatting SLAM. TopoGS-SLAM shows that once sufficient spatial coverage is established, accurate mapping is achievable under fixed topology, reformulating SLAM as a constant-dimensional optimization problem with bounded capacity that stabilizes scaling and prevents uncontrolled map growth. Experiments confirm that this stationary design preserves localization and reconstruction fidelity while improving memory efficiency and throughput, indicating that scalable Gaussian SLAM stems from controlled capacity rather than continual topology mutation.

\section*{Acknowledgments}
This work was supported by the National Science and Technology Council (NSTC), Taiwan, under Grant Nos. NSTC 115-2218-E-A49-010, NSTC 115-2221-E-A49-122-MY3, NSTC 115-2622-E-A49-006, NSTC 115-2622-8-002-009-TE2, NSTC 114-2634-F-002-004, NSTC 115-2634-F-007-003, NSTC 115-2634-F-002-012, and NSTC 115-2410-H-007-028. We also gratefully acknowledge the support provided by the MediaTek Advanced Research Center (MARC) through its industry--academia collaborative research program.

{
    \small
    \bibliographystyle{ieeenat_fullname}
    \bibliography{egbib}
}

\end{document}